\documentclass[9pt,conference]{IEEEtran}

\usepackage[colorlinks,urlcolor=blue,linkcolor=blue,citecolor=blue]{hyperref}

\usepackage{color,array}

\usepackage{cite}
\usepackage{graphicx}%
\usepackage{amsmath,amssymb,amsfonts}%
\usepackage{amsthm}%
\usepackage{mathrsfs}%
\usepackage{textcomp}%
\usepackage{manyfoot}%
\usepackage{algorithm}%
\usepackage{algorithmicx}%
\usepackage{algpseudocode}%
\usepackage{listings}%
 
\usepackage{mlmath}
\usepackage{mathtools}
\usepackage{subcaption}
\usepackage{enumitem}
\usepackage[table]{xcolor}
\usepackage{multirow} 
\usepackage{ulem}

\newtheorem{theorem}{Theorem}
\newtheorem{lemma}{Lemma}
\newtheorem{example}{Example}%
\newtheorem{remark}{Remark}%

\newtheorem{assumption}{Assumption}

\begin{document}
\title{Bridging the Gap between Total and Discounted Return Maximization in Reinforcement Learning}
\author{\IEEEauthorblockN{
        Shuyu Yin\IEEEauthorrefmark{1}, 
        Fei Wen\IEEEauthorrefmark{1}, 
        Peilin Liu\IEEEauthorrefmark{1},
        and Tao Luo\IEEEauthorrefmark{2}\IEEEauthorrefmark{3}
    }
    \IEEEauthorblockA{
        \IEEEauthorrefmark{1} Department of Electronic Engineering, Shanghai Jiao Tong University.\\
        \IEEEauthorrefmark{2} School of Mathematical Sciences, Institute of Natural Sciences, MOE-LSC, CMA-Shanghai, Shanghai Jiao Tong University.\\
        \IEEEauthorrefmark{3} Shanghai Artificial Intelligence Laboratory.}
}
\maketitle
\begin{abstract}
The optimal objective is a fundamental aspect of reinforcement learning (RL), as it determines how policies are evaluated and optimized. While total return maximization is the ideal objective in RL, discounted return maximization is the practical objective due to its stability. This can lead to a misalignment of objectives. 
To better understand the problem, we theoretically analyze the performance gap between the policy maximizes the total return and the policy maximizes the discounted return. Our analysis reveals that increasing the discount factor can be ineffective at eliminating this gap when environment contains cyclic states—a frequent scenario.
To address this issue, we propose two alternative approaches to align the objectives. The first approach achieves alignment by modifying the terminal state value, treating it as a tunable hyper-parameter with its suitable range defined through theoretical analysis. The second approach focuses on calibrating the reward data in trajectories, enabling alignment in practical Deep RL applications using off-policy algorithms. This method enhances robustness to the discount factor and improve performance when the trajectory length is large.
Our proposed methods demonstrate that adjusting reward data can achieve alignment, providing an insight that can be leveraged to design new optimization objectives to fundamentally enhance the performance of RL algorithms. The code is available at 
\href{https://github.com/dayhost/objective_alignment_RL}{here}.
\end{abstract}

\section{Introduction}

\begin{figure}[h]
\centering
\includegraphics[width=0.23\textwidth]{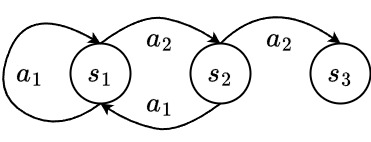}
\caption{A simple deterministic MDP with cyclic states.}
\label{fig::MDPExample}
\end{figure}

Reinforcement learning (RL) is a critical field in artificial intelligence and serves as a fundamental tool in various practical domains, such as gaming~\cite{mnih2015human, silver2016mastering}, unmanned autonomous vehicle control~\cite{liu2018energy, liu2024leader}, recommendation systems~\cite{zhao2021dear, afsar2022reinforcement}, robotics~\cite{franceschetti2021robotic, raffin2022smooth}, and quantitative trading~\cite{sun2022deepscalper, liu2020adaptive}. In these applications, the environment is typically modeled as a Markov Decision Process (MDP), with the objective of finding a policy that maximizes the total return, or equivalently minimizes the Bellman loss. However, directly optimizing the Bellman loss can lead to instability due to its lack of a contraction mapping property~\cite{sutton2018reinforcement}, and may result in divergence, particularly in scenarios with cyclic states, i.e., $s_1$ and $s_2$ in Figure \ref{fig::MDPExample}. To address these issues, the discounted return is widely adopted as a proxy for total return in deep RL (DRL) algorithms, like DQN~\cite{mnih2015human}, TRPO~\cite{schulman2015trust}, DDPG~\cite{lillicrap2015continuous}, TD3~\cite{fujimoto2018addressing}, and SAC~\cite{haarnoja2018soft}.


Despite its practicality, utilizing the discounted return as a proxy for the total return can lead to inconsistencies. It can result in a policy that maximize the discounted return fail to maximize the total return, a phenomenon we refer to as the \textbf{objective misalignment problem}. This discrepancy can be particularly pronounced: a policy that maximizes the discounted return can, paradoxically, minimize the total return. As shown in Example \ref{exam::simpleMDPexam} and Table \ref{tab::zeroTerminalStateValue}, 
the policy $\pi(s_1)=a_2, \pi(s_2)=a_2$ (highlighted in yellow) maximizes the discounted returns result in minimizing the total returns regardless of how close the discount factor is to one.


\begin{example}[MDP with cyclic states]
    \label{exam::simpleMDPexam}
    Consider a deterministic MDP with three states, \( S = \{s_1, s_2, s_3\} \), where \( s_3 \) is the terminal state. States \( s_1 \) and \( s_2 \) each have two available actions, \( A = \{a_1, a_2\} \). The transition dynamics are illustrated in Figure \ref{fig::MDPExample}. With a discount factor \( \gamma \in (0, 1) \), the reward function is defined as follows: \( r(s_1, a_1, s_1) = -1 \), \( r(s_1, a_2, s_2) = -1 \), \( r(s_2, a_1, s_1) = \gamma \), and \( r(s_2, a_2, s_3) = -1 \).
\end{example}

\begin{table}[htb]  
    \centering  
    \begin{tabular}{||>{\centering\arraybackslash}m{0.6cm} >{\centering\arraybackslash}m{0.6cm}|  
                    >{\centering\arraybackslash}m{1.5cm} >{\centering\arraybackslash}m{1.5cm}|  
                    >{\centering\arraybackslash}m{0.95cm} >{\centering\arraybackslash}m{0.95cm}||} 
         \hline   
         $\pi(s_1)$ & $\pi(s_2)$ & $G(\tau(s_1))$ & $G(\tau(s_2))$ & $R(\tau(s_1))$ & $R(\tau(s_2))$ \\ 
         \hline  
         \hline  
         $a_1$ & $a_1$ & $-1/(1-\gamma)$ & $-\gamma^2/(1-\gamma)$ & $-\infty$ & $-\infty$ \\ 
         \hline  
         $a_1$ & $a_2$ & $-1/(1-\gamma)$ & $-1$ & $-\infty$ & $-1$ \\
         \hline  
         \rowcolor{yellow}
         $a_2$ & $a_1$ & $-1$ & $0$ & $-\infty$ & $-\infty$ \\   
         \hline  
         $a_2$ & $a_2$ & $-1-\gamma$ & $-1$ & $-2$ & $-1$ \\
         \hline  
    \end{tabular}  
    \caption{
    The optimal policy for maximizing the discounted return $G$ can paradoxically minimize the total return $R$. The terminal state value is set to zero and \(\gamma \in (0, 1)\). $\tau(s)$ denote the trajectory start from $s$. The policy \( \pi(s_1) = a_2, \pi(s_2) = a_1 \) (highlighted in yellow) maximizes the discounted return while simultaneously minimizing the total return.
    }  
    \label{tab::zeroTerminalStateValue}  
\end{table}

\begin{figure}[h]
\centering
\includegraphics[width=0.45\textwidth]{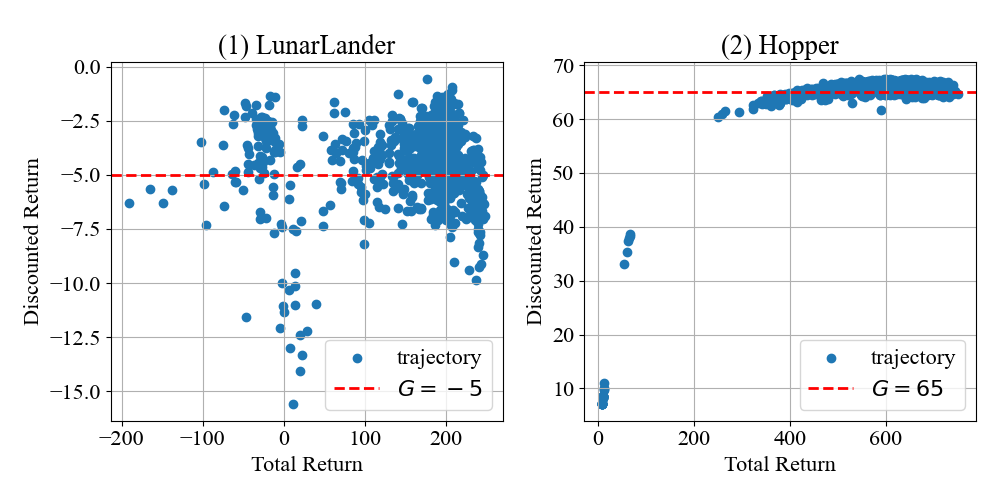}
\caption{
Demonstration of the objective misalignment problem. \(G\) represents the discounted return. Each point corresponds to a trajectory collected from the environment, with discounted returns calculated using a discount factor of \(0.97\). As shown, for a given discounted return, the total return varies widely. 
}
\label{fig::objective_misalignment_problem}
\end{figure}

Specifically, for practical DRL applications, this problem can be described as follow: the discounted returns of trajectories do not increase monotonically with their total returns. For example, in Figure \ref{fig::objective_misalignment_problem}, each point represents a trajectory collected from the environment. In both subfigures, trajectories with the same discounted returns can exhibit significant variation in their total returns. This phenomenon restricts the algorithm's capacity to exploit high-quality trajectory data and to explore superior trajectories, ultimately degrading performance. This is also one of the causes behind the regularization effect of the discount factor~\cite{amit2020discount, hu2022role}.

Above phenomena underscore the importance of addressing the objective misalignment problem. To further investigate this issue, we analyzed the performance gap—specifically, the suboptimality bound—between the policy derived from minimizing the discounted Bellman loss, which is equivalent to maximizing the discounted return, and the policy that maximizes the total return. A smaller suboptimality bound indicates better performance and closer alignment with the true objective.
As stated in Theorem \ref{thm::valueSuboptimality}, this bound can be decomposed into three terms: \(O\left(L_{\max} \fL(f,\gamma)^{1/2}\right)\), \(O\left(L_{\max} (1-\gamma)^{1/2} \fL(f,\gamma)^{1/4}\right)\), and \(O\left(L_{\max}(1-\gamma)\right)\), where \(L_{\max}\) is the maximum trajectory length, and \(\fL(f, \gamma)\) is the discounted Bellman loss. The first two terms can be reduced by minimizing \(\fL(f, \gamma)\). However, the third term, determined solely by \(\gamma\) and \(L_{\max}\), is unaffected by minimizing the loss. This yields a key insight: \textbf{minimizing the discounted Bellman loss does not guarantee to improve performance in terms of total return.} This insight is validated by Example \ref{exam::simpleMDPexam} and Table \ref{tab::zeroTerminalStateValue}.

A common solution to this issue is to increase \(\gamma\) close to one. However, this approach introduces new challenges: (1) it destabilizes training~\cite{gao2022partial, wang2024fractal}; (2) it slows convergence~\cite{amit2020discount}; and (3) it can become ineffective in MDPs with cyclic states (\(L_{\max} = \infty\)), as illustrated in Example \ref{exam::simpleMDPexam} and the suboptimality analysis above. Alternative methods, such as redefining state values for cyclic states~\cite{gao2022partial}, aim to improve stability when the discount factor is near one. However, they may also fail in environments where the optimal policy necessitates constructing cyclic states, such as in CartPole. Therefore, it is essential to explore alternative methods to align the two objectives instead of relying solely on increasing $\gamma$.

In this work, we propose two approaches to achieve objective alignment. First, we adjust the discounted returns by modifying the terminal state value, enabling policies that maximize the modified discounted returns to align with those maximizing the original total returns. By changing the terminal state value, the comparative relationship of discounted returns between two trajectories can be reversed, while the relationship of total returns remains unaffected. In Table \ref{tab::nonzeroTerminalStateValue}, we demonstrate the effectiveness of terminal state value modification. We highlight that \textbf{the terminal state value can be treated as a tunable hyper-parameter to address the objective misalignment problem, and a carefully designed value has the potential to enhance performance.} Building on this insight, we derive sufficient conditions for aligning the optimal policies for two objectives theoretically, and validate them experimentally.

To extend our insights to boarder DRL applications, we introduce a trajectory reward data calibration method to modify discounted returns. This approach ensures a monotonic increase of the modified discounted returns with respect to original total returns across sampled trajectories. Unlike terminal state value modification, this method distributes the required alignment adjustments across the reward data throughout the trajectory, which improves training stability. Our method can seamlessly integrates with off-policy DRL algorithms. It enhances the robustness of these algorithms to the discount factor and improves performance across tasks with long trajectory lengths. The effectiveness of our method further highlights that \textbf{the reward data calibration technique can be leveraged to design new proxies that inherently increases monotonically with the total return, thereby fundamentally improving the performance of all RL algorithms.}

\textbf{Contributions:}
This work makes three key contributions: 

(1) \textbf{Identification and Analysis of the Objective Misalignment Problem.} 
To the best of our knowledge, we are the first to explicitly identify and systematically analyze the objective misalignment problem, a fundamental issue in RL. Through concrete examples and theoretical analysis, we demonstrate how this misalignment can significantly degrade the performance of RL algorithms and show that simply increasing the discount factor cannot resolve the problem.

(2) \textbf{Policy Alignment via Terminal State Value Modification.}
We propose a novel approach to align policies that maximize both the total return and the discounted return by modifying the terminal state value, and we derive suitable ranges for it through theoretical analysis. This highlights that the terminal state value can serve as a tunable hyper-parameter to enhance the performance of RL algorithms.

(3) \textbf{Objective Alignment via Trajectory Reward Data Calibration.} 
To address objective misalignment in broader DRL applications, we propose a trajectory reward data calibration method. This approach enhances the robustness of off-policy algorithms to the discount factor and improves performance across tasks with large trajectory length. Additionally, our method provides valuable insights for designing new proxies for total return.

\section{Background}

\subsection{Related works}

This work addresses the issue of objective misalignment. Existing studies focus on aligning the two objectives by increasing the discount factor, such as the work of Gao \textit{et al.}~\cite{gao2022partial}. This work prevents agents from becoming trapped in cyclic states, thereby avoiding instability caused by large discount factors. By allowing the use of discount factors close to one, their method ensures that the discounted return closely approximates the total return. However, this approach fails when the optimal policy inherently requires the construction of cyclic states, as seen in the CartPole problem. In contrast, our work proposes alternative approaches to align the two objectives, rather than simply increasing the discount factor.

Establishing relationships between different objectives in RL has attracted attention since the early stages of RL research. For instance, Mahadevan \textit{et al.}~\cite{mahadevan1996optimality} studied various criteria for policy optimality and established conversion relationships between them. Other works have compared policies derived from maximizing discounted return versus maximizing average return. Additionally, Mahadevan \textit{et al.}~\cite{mahadevan1994discount} numerically compared the performance of policies learned through maximizing discounted return and average return, concluding that maximizing discounted return yields superior policies. Kakade \textit{et al.}~\cite{kakade2001optimizing} analyzed the performance of average return optimization through discounted Bellman loss, while Tsitsiklis \textit{et al.}~\cite{tsitsiklis2002average} examined the performance gap between these two approaches within the temporal-difference learning framework. More recently, studies by Siddique \textit{et al.}~\cite{siddique2020learning} and Dewanto \textit{et al.}~\cite{dewanto2022examining} have extended this comparison to DRL, analyzing suboptimality and experimentally evaluating the advantages and limitations of both approaches across different tasks. In this work, we aim to contribute to this field by bridging the gap between two commonly used objectives.

The analysis of performance gap in this work is closely related to the field of suboptimality bound analysis in RL. There are already numerous studies that analyze the suboptimality bound within a single objective framework. Antos \textit{et al.}~\cite{antos2008learning} defined the optimality objective in terms of discounted return and analyzed the suboptimaility bound derived from a single trajectory as training data. Fan \textit{et al.}~\cite{fan2020theoretical} focused on maximizing discounted return, examining the suboptimaility bound of policies obtained through the Deep Q-Learning algorithm. In contrast, Duan \textit{et al.}~\cite{duan2021risk} utilized total return as the optimality objective, deriving suboptimality bounds for policies obtained by minimizing the Bellman loss under both single and double sampling scenarios. Xie \textit{et al.}~\cite{xie2020q} aimed to maximize discounted return as well, proposing algorithms that optimize variations of the discounted Bellman loss while analyzing their associated suboptimality bounds. Similarly, Long \textit{et al.}~\cite{long2021an} studied the suboptimality bounds of policies learned by minimizing the Bellman loss, using total return as the optimality objective and employing two-layer neural networks. While these studies primarily examine suboptimality within a single objective framework, our work explores the performance gap across two distinct objectives.

\subsection{Preliminaries}

A MDP is defined by the tuple $(S, \partial S, A, P, r, \gamma)$. $S$ is the \textbf{state set}, with each $s \in S$ representing a state. $\partial S \subset S$ is the \textbf{terminal state set}, and $S^{\circ} = S \backslash \partial S$ is the \textbf{non-terminal state set}. $A$ is the \textbf{action set}, with each $a \in A$ representing an action. $P: S \times A \to \fP(s)$ is the \textbf{Markov transition kernel} and $P(\cdot|s,a)$ is the probability for next state. $r: S \times A \times S \to \mathbb{R}$ is the \textbf{reward function}. $\gamma \in (0,1)$ is the \textbf{discount factor}.  

A state $s \in S$ is called a \textbf{terminal state} if $P(s | s, a) = 1$ for all $a \in A$. A \textbf{policy} $\pi: S \to \fP(A)$ maps any $s \in S$ to a probability distribution $\pi(\cdot|s)$ over $A$, and $\Pi$ denotes the \textbf{policy set}, containing all possible policies. Additionally, the size of a set $\Omega$ is denoted as $|\Omega|$.  

The \textbf{transition kernel under a policy $\pi$} is denoted as $P^{\pi}: S \to \fP(S)$, where $P^{\pi}(s'|s) = \int_A \pi(a|s) P(s'| s, a) \, da$. For simplicity, we omit explicit dependence on $\pi$ in the following definitions: A \textbf{walk} $\{s_0, \ldots, s_{L-1}\}$ is a sequence of states such that $P^{\pi}(s_{i+1}|s_{i}) > 0$ for $i = 0, \ldots, L-1$. The \textbf{length} of the walk is $L$. A \textbf{path} is a walk where no states are repeated. A \textbf{trajectory} $\tau = \{s_0, a_0, s_1, \ldots, s_{L-1}, a_{L-1}, s_L\}$ combines a walk with its corresponding actions, and $s_L \in \partial S$ if $L < \infty$. The maximum trajectory length is denoted as $L_{\max}$. Moreover, a state $s' \in S$ is \textbf{accessible} from $s \in S$ if there exists a walk from $s$ to $s'$.  

The \textbf{$k$-step transition kernel} $P_k^{\pi}: S \to \fP(S)$ is the extension of the transition kernel under a policy. The \textbf{$k$-step transition probability} from $s \in S$ to $s' \in S$ in $k$ steps, $P_k^{\pi}(s'|s)$, is defined as:
\begin{equation*}
    \begin{aligned}
        & P^{\pi}_k(s'|s) =  \int_{S} P^{\pi}(s_1|s) \cdots \int_{S} P^{\pi}(s_{k-2}|s_{k-3}) \\
        &~~~~ \int_{S} P^{\pi}(s_{k-1}|s_{k-2}) P^{\pi}(s'|s_{k-1}) \, ds_{k-1} \cdots ds_1.
    \end{aligned}
\end{equation*}
If $s' \in \partial S$, $P^{\pi}_k(s'|s)$ is referred to as the \textbf{$k$-step absorption probability}.  

Given a trajectory $\tau$ with length $L$, the \textbf{discounted return $G(\tau)$ and total return $R(\tau)$ for trajectory $\tau$} is defined as 
\begin{equation*}
    G(\tau) = \sum_{t=0}^{L-1} \gamma^t r(s_t,a_t,s_{t+1}), \,\, R(\tau) = \sum_{t=0}^{L-1}r(s_t,a_t,s_{t+1}).
\end{equation*}
The \textbf{expected total return} or \textbf{state value} are defined as: 
\begin{equation*}
    \begin{aligned}
        V_{1}^{\pi}(s) & = \mathbb{E}_{\pi, P} \left[\sum_{t=0}^{\infty} r(s_t,a_t,s_{t+1}) \Bigg| s_0=s \right] \\
        & = \int_{S} \sum_{k=0}^{\infty} P^{\pi}_{k}(s'|s) r^{\pi}_{P}(s') \, ds',
    \end{aligned}
\end{equation*}
where $r_{P}^{\pi}(s') = \int_S \int_A \pi(a | s') P(s'' | s', a) r(s', a, s'') \, da \, ds''$. For simplicity, we write $V^{\pi}(s)$ to denote $V_1^{\pi}(s)$.
The \textbf{expected discounted return} or \textbf{discounted state value} are:  
\begin{equation*}
    \begin{aligned}
        V^{\pi}_{\gamma}(s) & = \mathbb{E}_{\pi,P} \left[\sum_{t=0}^{\infty} \gamma^t r(s_t,a_t,s_{t+1}) \Bigg| s_0=s \right] \\
        & = \int_{S} \sum_{k=0}^{\infty} \gamma^k P^{\pi}_k(s'|s) r^{\pi}_{P}(s') \, ds'.
    \end{aligned}
\end{equation*}
The corresponding \textbf{discounted state-action value} is:  
\begin{equation*}
    \begin{aligned}
        Q^{\pi}_{\gamma}(s, a) = \mathbb{E}_{\pi, P} \left[\sum_{t=0}^{\infty} \gamma^t r(s_t,a_t,s_{t+1}) \Bigg| s_0 = s, a_0 = a \right].
    \end{aligned}
\end{equation*}

The \textbf{optimal policy} that maximizes total return is denoted by $\pi^+$, and $\pi_\gamma^+$ denotes the optimal policy for discounted return. Similarly, $\pi^-$ and $\pi_\gamma^-$ denote the \textbf{worst policies} that minimize total and discounted returns, respectively.  

For discrete action spaces, the \textbf{discounted Bellman equation} for $(s, a) \in S \times A$ is:  
\begin{equation*}
    Q(s,a) = \fT Q(s,a) : = \mathbb{E}_{P} \left[ r(s,a,s') + \gamma  \max_{a'} Q(s', a') \right],
\end{equation*}
where $\fT$ is the \textbf{Bellman operator}. Specifically, when $s' \in \partial S$, the $\max_{a'} Q(s', a')$ reduces to a fixed terminal state value. Solving the discounted Bellman equation can obtain $\pi_\gamma^+$, and the solution can be obtained by minimizing the \textbf{discounted Bellman (optimal) loss}:  
\begin{equation}
    \label{eq::BellmanLossDefi}
        \fL(Q, \gamma) = \mathbb{E}_{s \sim \mu, a \sim \nu} \Big[ Q(s,a) -  \fT Q(s,a) \Big]^2,
\end{equation} 
where $\gamma \in (0,1)$, $\mu$ is a distribution over $S^{\circ}$, and $\nu$ is a distribution over $A$. When $\gamma = 1$, the loss becomes the \textbf{Bellman (optimal) loss}, denoted $\fL(Q, 1)$. The solution to the discounted Bellman equation is denoted as $Q_\gamma^*$.  

We now state the assumptions used in this work:  

\begin{assumption}[bounded reward function]  
\label{assum::rewardBound}  
For all $(s, a, s') \in S \times A \times S$, $|r(s, a, s')| < \infty$. Specifically:  1) for all $(s, a, s') \in S^{\circ} \times A \times S$, there exist $r_{\min}$ and $r_{\max}$ such that $r_{\min} \leq r(s, a, s') \leq r_{\max}$, 2) for all $s \in \partial S$ and $a \in A$, $r(s, a, s) = (1-\gamma)\fV_T$, where $\fV_T \in \mathbb{R}$ is the terminal state value.  
\end{assumption}  

\begin{remark}  
    This assumption guarantees the boundedness of the reward function, a standard setting in RL. For a terminal state $s \in \partial S$, by setting $r(s, a, s) = (1-\gamma) \fV_T$, its value converges to $\fV_T$. For any policy $\pi \in \Pi$, the expected reward satisfies $r_{\min} < r_{P}^{\pi}(s) < r_{\max}$. In the case of discrete action space $A$, the existence of the optimal Q-function $Q_\gamma^*$ follows from the contraction mapping theorem~\cite{sutton2018reinforcement}.
\end{remark}

\section{Analysis of Objective Misalignment Problem}
\label{sec::theoreticalAnalysis}

From Example \ref{exam::simpleMDPexam}, we observe that when the objectives are misaligned, even if the discounted state values are estimated accurately (i.e., the discounted Bellman loss is zero), the optimal policies under the two objectives can still differ significantly. To formally analyze this discrepancy, we introduce the following assumption:

\begin{assumption}[concentrability] 
    \label{assum::concentrability}
    Let $\rho^{\pi}_{s_0}$ denote the distribution over $S \times A$ starting from state $s_0 \in S$ and following a policy $\pi \in \Pi$, and let $\kappa$ denote a distribution over $S^{\circ} \times A$ where $\kappa(s,a)=\mu(s) \cdot \nu(a)$. There exists a $\xi \in \mathbb{R}$ such that 
    \begin{equation*}
        \sup_{(s,a) \in S \times A} \frac{d \rho^{\pi}_{s_0}}{d \kappa}(s,a) \leq \xi, \quad \text{for all } \pi \in \Pi.
    \end{equation*}
\end{assumption}

This assumption ensures that the sample dataset is sufficiently large to provide adequate coverage for any given policy. It is a standard assumption in suboptimality analysis~\cite{antos2008learning, duan2021risk}. Additionally, to simplify the analysis process, we define a function space of bounded functions as follows:
\begin{equation*}
    \fF_{M} = \left\{f: S \times A \to \mathbb{R} \, \big| \, |f| \leq M \text{ for all } (s,a) \in S \times A \right\}.
\end{equation*}
Our analysis can be easily extended to a neural network setting without the need for explicit value bounds.

We now present the performance gap-suboptimality bound-between the policy obtained by minimizing the discounted Bellman loss and the policy that maximizes total return. We define the maximum trajectory length as $L_{\max} \in \mathbb{N} \cup \{ +\infty\}$, indicating $\int_{\partial S} P_{L_{\max}}^{\pi}(s'|s) ds'=1$ for any $s \in S$ and $\pi \in \Pi$.

\begin{theorem}[suboptimality]
    \label{thm::valueSuboptimality}
    Consider a stochastic MDP with a discrete action space of size $|A| < \infty$ and a maximum trajectory length $L_{\max}$, satisfying Assumptions \ref{assum::rewardBound}, \ref{assum::concentrability}, and $\fV_T=0$. Then for any $f \in \fF_M$, we have:
    \begin{equation*}
        \begin{aligned}
            \| V^{\pi^+} - V^{\pi_f} \|_{1,\mu} & \leq  2 L_{\max} \left( \xi (1-\gamma) M \right)^{\frac{1}{2}} \left( \fL(f,\gamma)\right)^{\frac{1}{4}} \\
            & + 2 L_{\max} \left(\xi \fL(f,\gamma) \right)^{\frac{1}{2}} +  2L_{\max} \xi^{\frac{1}{2}} (1-\gamma) M,
        \end{aligned}
    \end{equation*}
    where $\pi_f$ is the policy derived from $f$.
\end{theorem}

\begin{proof}
    We first relate the approximation error of $\fL(f, 1)$ to $\fL(f, \gamma)$. By the Taylor expansion, we have:
    \begin{equation}
        \label{eq::lossTaylorExpansion}
        \begin{aligned}
            \fL(f, 1) - \fL(f, \gamma) & = \nabla_{\gamma} \fL(f, \gamma)(1 - \gamma) \\
            & ~~~~ + \frac{1}{2} \nabla_{\gamma}^2 \fL(f, \gamma) (1 - \gamma)^2.
        \end{aligned}
    \end{equation}
    From the definition of Bellman loss \eqref{eq::BellmanLossDefi}, the first and second derivative of $\fL(f, \gamma)$ respect to $\gamma$ are:
    \begin{equation}
        \label{eq::gradOfLoss}
        \begin{aligned}
            \nabla_{\gamma} \fL(f, \gamma) & = \mathbb{E} \bigg[ \max_{a'\in A} f(s',a') \Big(f(s,a) - \fT f(s,a) \Big) \bigg],
        \end{aligned}
    \end{equation}
    \begin{equation}
        \label{eq::hessianOfLoss}
        \nabla_{\gamma}^2 \fL(f, \gamma) = \mathbb{E}\bigg[\max_{a' \in A}f(s',a')\bigg]^2.
    \end{equation}
    Combining \eqref{eq::lossTaylorExpansion}, \eqref{eq::gradOfLoss}, \eqref{eq::hessianOfLoss} and the definition of $\fF_{M}$, we can bound $\fL(f, 1) - \fL(f, \gamma)$ as:
    \begin{equation}
        \label{eq::discountedLossDiff}
        \begin{aligned}
            \fL(f, 1) - \fL(f, \gamma) \leq (1 - \gamma) M \sqrt{\fL(f, \gamma)} + \frac{1}{2} (1 - \gamma)^2 M^2.
        \end{aligned}
    \end{equation}

    To relate suboptimality to Bellman loss, we apply a telescoping sum argument similar to \cite{duan2021risk} (Lemma 3.2). For each $s \in S$, we have:
    \begin{equation*}
        V^{\pi^+}(s) - V^{\pi_f}(s) \leq 2 L_{\max} \sqrt{\xi \cdot \fL(f, 1)}.
    \end{equation*}
    Since $V^{\pi^+}(s) \geq V^{\pi_f}(s)$ for all $s \in S$, taking the expectation over $s \in \mu$ yields:
    \begin{equation}
        \label{eq::suboptimalityToBellmanLoss}
        \| V^{\pi^+} - V^{\pi_f} \|_{1,\mu} \leq 2 L_{\max} \sqrt{\xi \cdot \fL(f, 1)}.
    \end{equation}
    Combining \eqref{eq::discountedLossDiff} and \eqref{eq::suboptimalityToBellmanLoss}, we have:
    \begin{equation*}
        \begin{aligned}
            \| V^{\pi^+} - V^{\pi_f} \|_{1,\mu} & \leq 2 L_{\max} \sqrt{\xi} \Big( \fL(f, \gamma) + (1 - \gamma) M \sqrt{\fL(f, \gamma)} \\
            &~~~ + \frac{1}{2} (1 - \gamma)^2 M^2 \Big)^{1/2}.
        \end{aligned}
    \end{equation*}
    After applying Jensen's inequality, we split the term inside the square root into three parts and get the final result.
\end{proof}

The result in Theorem \ref{thm::valueSuboptimality} comprises three terms. The first two are related to the discounted Bellman loss and vanish as the loss approaches zero. In contrast, the third term depends solely on the discount factor $\gamma$ and the maximum trajectory length $L_{\max}$, which are intrinsic properties of the environment. This implies that the third term cannot be further minimized through optimization. \textbf{When the third term dominates the performance gap, minimizing the discounted Bellman loss does not necessarily lead to improved performance.}  

This insight is supported by Example \ref{exam::simpleMDPexam} and Table \ref{tab::zeroTerminalStateValue}. Example \ref{exam::simpleMDPexam} demonstrates that, regardless of how close the discount factor $\gamma$ is to one, in environments with cyclic states ($L_{\max} = \infty$), optimizing the discounted return may fail to improve performance with respect to the total return. As cyclic states are common in practical RL environments, \textbf{this analysis highlights the need for alternative approaches to address the objective misalignment problem, rather than relying solely on increasing the discount factor.}

\section{Alignment Condition Analysis}
\label{sec::policyAlignment}

Based on the analysis of Theorem \ref{thm::valueSuboptimality}, we propose addressing the objective misalignment problem by directly modifying the discounted returns of trajectories. Further examination of Example \ref{exam::simpleMDPexam} demonstrates that modifying the terminal state value effectively aligns the policy that maximizes discounted returns with the policy that maximizes total returns. Specifically, setting the terminal state value to \( V(s_3) = 2/\gamma \) produces the updated values in Table \ref{tab::nonzeroTerminalStateValue}. Under this adjustment, the optimal policy based on discounted returns becomes \( \pi(s_1) = a_2,\) \(\pi(s_2) = a_2\), which aligns with the policy that maximizes total returns (highlighted in yellow). Notably, adjusting the terminal state value only alters the comparative relation of discounted returns while preserving the relation between total returns.

\begin{table}[htb]
    \begin{tabular}{||>{\centering\arraybackslash}m{0.6cm} >{\centering\arraybackslash}m{0.6cm}|  
                    >{\centering\arraybackslash}m{1.5cm} >{\centering\arraybackslash}m{1.5cm}|  
                    >{\centering\arraybackslash}m{0.95cm} >{\centering\arraybackslash}m{0.95cm}||} 
         \hline 
         $\pi(s_1)$ & $\pi(s_2)$ & $G(\tau(s_1))$ & $G(\tau(s_2))$ & $R(\tau(s_1))$ & $R(\tau(s_2))$ \\ 
         \hline
         \hline
         $a_1$ & $a_1$ & $-1/(1-\gamma)$ & $-\gamma^2 / (1-\gamma)$ & $-\infty$ & $-\infty$ \\ 
         \hline
         $a_1$ & $a_2$ & $-1/(1-\gamma)$ & $1$ & $-\infty$ & $-1$ \\
         \hline
         $a_2$ & $a_1$ & $-1$ & $0$ & $-\infty$ & $-\infty$ \\ 
         \hline
         \rowcolor{yellow}
         $a_2$ & $a_2$ & $-1+\gamma$ & $1$ & $-2$ & $-1$ \\
         \hline
    \end{tabular}
    \caption{
    The optimal policy for the discounted return $G$ can be aligned with the optimal policy for the total return $R$ by adjusting the terminal state value. $\tau(s)$ denote the trajectory start from $s$. When the terminal state value is set to \( 2/\gamma \), the policy \( \pi(s_1) = a_2 \), \( \pi(s_2) = a_2 \) (highlighted in yellow) maximizes both the discounted return and the total return.
    }
    \label{tab::nonzeroTerminalStateValue}
\end{table}

The core idea of aligning the two objectives by adjusting the terminal state value is to ensure that, for any two given trajectories, \(\tau_1\) and \(\tau_2\), the relationship between their discounted returns and total returns remains consistent. Notably, given two trajectories differ in length, it is always possible to identify a terminal state value that aligns the two objectives for these trajectories. This concept is formalized in the following lemma. As the lemma is straightforward, we omit its proof.

\begin{lemma}[terminal state values reverse relation]
\label{lem::terminalReverseRelation}
Given two trajectories \(\tau_1 = \{s^1_0, a^1_0, \ldots, s^1_{L_1}\}\) and \(\tau_2 = \{s^2_0, a^2_0, \ldots, s^2_{L_2}\}\) with lengths \(L_1\) and \(L_2\), respectively. If $L_1,L_2 < \infty$, \(L_1 \neq L_2\) and \(\sum_{i=0}^{L_1 - 1} r(s^1_i, a^1_i, s^1_{i+1}) > \sum_{j=0}^{L_2 - 1} r(s^2_j, a^2_j, s^2_{j+1})\), then for any \(\gamma \in (0, 1)\), there exists a \(\fV_T \in \mathbb{R}\) such that $\sum_{i=0}^{L_1 - 1} \gamma^{i} r(s^1_i, a^1_i, s^1_{i+1}) + \gamma^{L_1} \fV_T < \sum_{j=0}^{L_2 - 1} \gamma^j r(s^2_j, a^2_j, s^2_{j+1}) + \gamma^{L_2} \fV_T.$
\end{lemma}

We now proceed to derive the sufficient conditions for aligning the two objectives under three general cases: (1) a positive reward function, (2) a negative reward function, and (3) a constant reward function. For each case, we identify the range of terminal state values that can achieve alignment between the two objectives. To facilitate this analysis, we first introduce an assumption regarding the accessibility of terminal states in the MDP environment.

\begin{assumption}[accessibility of terminal states]  
\label{assum::terminalAccessibility}  
Given a terminal state set \(\partial S \neq \emptyset\), we assume two conditions hold: 
\begin{enumerate}[label = (\arabic*), itemindent = 0pt, labelindent = 0pt, labelwidth = 2em, labelsep = 5pt, leftmargin = *, itemsep=2pt, topsep=0pt, parsep=0pt]  
    \item There exist \(s \in S^{\circ}\), \(s' \in \partial S\), and \(\pi \in \Pi\) such that \(s'\) is accessible from \(s\) under the transition kernel \(P^{\pi}\).  
    \item There exists a policy \(\pi \in \Pi\) such that for every \(s \in S^{\circ}\), \(P^{\pi}(s'| s) = 0\) for all \(s' \in \partial S\).  
\end{enumerate}  
\end{assumption}  

Assumption \ref{assum::terminalAccessibility} establishes the conditions for whether a non-terminal state can reach a terminal state under a given policy. Condition (1) ensures the existence of a policy under which at least one non-terminal state can transition to a terminal state, while condition (2) ensures the existence of a policy under which no non-terminal state reaches a terminal state. This assumption is crucial for deriving the explicit forms of the optimal or worst policies when the reward function is positive or negative, respectively.

To proceed, we first present two lemmas. Lemma \ref{lem::walkLength} establishes the maximum distance between two accessible states, while Lemma \ref{lem::absorpProb} demonstrates that the probability of reaching a terminal state increases with step size.

\begin{lemma}[maximum access distance]
    \label{lem::walkLength}
    In a stochastic MDP with a discrete state space of size $|S| < \infty$ and a policy $\pi \in \Pi$, if a state $s' \in S$ is accessible from $s \in S$, where $s \neq s'$, then there exists a $k \leq |S| - 1$ such that $P_k^{\pi}(s'|s) > 0$.
\end{lemma}

\begin{proof}
    Since $s'$ is accessible from $s$, there exists a policy $\pi \in \Pi$ that can construct a walk from $s$ to $s'$, i.e., a sequence $s_0,s_1,s_2,...,s_{L-1}$ where $s_0=s$, $s_{L-1}=s'$ and $P^{\pi}(s_{i+1}|s_i)>0$ for $0 \leq i < L$.

    To construct a path (a walk without repeating states), we can remove all repeated states from the sequence. If $s_i = s_{i+j}$ for some $i$ and $j>0$, we remove all intermediate states $s_i,...,s_{i+j-1}$. Repeating this procedure eliminates all cyclic states, resulting in a path of length $N$, where $N \leq |S|$, because a path cannot contain more states than the state space.

    Let $k = N-1$, which is the number of transitions in the path. From the definition of $k$-step transition probability, we have:
    \begin{equation*}
        \begin{aligned}
            P^{\pi}_k(s'|s) \geq \prod_{i=1}^{N-1} P^{\pi}(s_{i}|s_{i-1}) > 0,
        \end{aligned}
    \end{equation*}
    where $s_0 = s$ and $s_{N-1}=s'$. Since $k = N-1 \leq |S|-1$, the statement is proved.
\end{proof}

\begin{lemma}[monotonicity of absorption probability]
    \label{lem::absorpProb}
    In a stochastic MDP with compact state space $S \subset \mathbb{R}^d$, action space $A$, and policy $\pi \in \Pi$, for any non-terminal state $s \in S^{\circ}$, terminal state $s' \in \partial S$, and any $k \geq 1$, the $k$-step absorption probability satisfies $P^{\pi}_{k+1}(s'|s) \geq P^{\pi}_k(s'|s)$.
\end{lemma}

\begin{proof}
    By the definition of $k+1$-step transition probability and the definition of terminal states, we have:
    \begin{equation*}
        \begin{aligned}
            & P^{\pi}_{k+1}(s'|s) = \int_{S} P^{\pi}_k(s''|s)P^{\pi}(s'| s'') \,d s'' \\
            \geq & P^{\pi}_{k}(s'|s)P^{\pi}(s'| s') = P^{\pi}_{k}(s'|s).
        \end{aligned}
    \end{equation*}
    Monotonicity of the absorption probability is established. 
\end{proof}

Given a stochastic MDP with a discrete state space of size \(|S|\), the minimum positive \((|S| - 1)\)-step absorption probability is defined as \(\delta = \min \Delta\), where \(\Delta\) is given by:  
\begin{equation*}
    \Delta := \bigcup_{\pi \in \Pi} \left\{ P^{\pi}_{|S|-1}(s'|s ) > 0 \;\middle|\; s \in S,\, s' \in \partial S \right\}.
\end{equation*}

\begin{remark}
    \label{remark::minAbsorptionProb}
    For a deterministic MDP that satisfies Assumption \ref{assum::terminalAccessibility}, we have \(\delta = 1\). In a deterministic environment, a non-terminal state either has access to a terminal state or does not, meaning the probability for a non-terminal state to reach a terminal state is either zero or one. Assumption \ref{assum::terminalAccessibility} (1) guarantees the existence of a policy that allows a non-terminal state to access a terminal state.
\end{remark} 

\begin{remark}
    From Lemma \ref{lem::walkLength}, if for a given state \(s \in S\) and policy \(\pi \in \Pi\), we have \(P^{\pi}_{|S|-1}(s'|s) = 0\) for all \(s' \in \partial S\), then \(s\) cannot reach any terminal state. Besides, under Assumption \ref{assum::terminalAccessibility} (1), we will have \(\Delta \neq \emptyset\). Furthermore, from Lemma \ref{lem::absorpProb}, if \(P^{\pi}_k(s'|s) > 0\) for some \(s \in S\), \(s' \in \partial S\), \(\pi \in \Pi\), and \(k \geq |S|\), then \(P^{\pi}_k(s'|s) \geq \delta\).
\end{remark}

With these results in hand, we now proceed to establish sufficient conditions for the alignment.

\begin{theorem}[align optimal policies with positive rewards]
    \label{thm::condition4Optimal}
    For a stochastic MDP with a discrete state space of size $|S| < \infty$, satisfying Assumptions \ref{assum::rewardBound} and \ref{assum::terminalAccessibility}, and $r_{\min} > 0$, if we set
    \begin{equation*}
        \fV_T < \frac{r_{\min} - r_{\max}}{\delta \gamma^{|S|-1} (1-\gamma)},
    \end{equation*}
    then $\pi^{+}_{\gamma} \in \arg \max_{\pi} V^{\pi}$. This implies that the optimal policies of the two objectives are aligned.
\end{theorem}

\begin{proof}
    Since \(r(s, a, s') > 0\) for all \((s, a, s') \in S^{\circ} \times A \times S\), the optimal policy \(\pi^+\) for the total return should prevent all non-terminal states \(s \in S^{\circ}\) from accessing any terminal state \(s' \in \partial S\). This results in \(V^{\pi^+}(s) = +\infty\) for all \(s \in S^{\circ}\). The existence of the policy is guaranteed by Assumption \ref{assum::terminalAccessibility} (1).  
    
    To ensure that \(\pi_{\gamma}^+\) is consistent with \(\pi^+\), the discounted state values under \(\pi^+\) must be greater than the discounted state values under any other policy \(\pi\) that allows any \(s \in S^{\circ}\) to access any \(s' \in \partial S\). This condition can be expressed as the following inequality:
    \begin{equation}
        \label{eq::optimal_condition_inequality}
        I^{\pi^+}_1 > I^{\pi}_2 + I^{\pi}_3,
    \end{equation}
    where
    \begin{equation}
        \label{eq::rewardDefTerms}
        \begin{aligned}
            I^{\pi^+}_1 &:= \int_{S^{\circ}} \sum_{k=0}^{\infty} \gamma^k P^{\pi^+}_{k}(s'|s) r^{\pi^+}_{P}(s') ds',  \\
            I^{\pi}_2 &:= \int_{S^{\circ}} \sum_{k=0}^{\infty} \gamma^k P^{\pi}_k(s'|s)  r^{\pi}_{P}(s') ds', \\
            I^{\pi}_3 &:= \int_{\partial S} \sum_{k=0}^{\infty} \gamma^k P^{\pi}_k(s'|s) ds' \cdot (1 - \gamma) \fV_T. 
        \end{aligned}
    \end{equation}

    Since $\int_{S^{\circ}} P^{\pi^+}_k(s'|s) ds' = 1$ for all $k \geq 0$, $I^{\pi^+}_1$ satisfies:
    \begin{equation*}
        I^{\pi^+}_1 \geq \sum_{k=0}^{\infty} \gamma^k r_{\min} = \frac{r_{\min}}{1 - \gamma}.
    \end{equation*}

    For any other policy $\pi$, since $\int_{S^{\circ}} P^{\pi}_k(s'|s) ds' \leq 1$ for all $k \geq 0$ and $r_{\max} > 0$, $I^{\pi}_2$ satisfies:
    \begin{equation*}
        I^{\pi}_2 \leq \sum_{k=0}^{\infty} \gamma^k r_{\max} = \frac{r_{\max}}{1 - \gamma}.
    \end{equation*}

    For $I^{\pi}_3$, based on the definition of $\delta$ and using Lemmas \ref{lem::walkLength} and \ref{lem::absorpProb}, we have:
    \begin{equation*}
        \begin{aligned}
            I^{\pi}_3 \geq & (1 - \gamma) \sum_{k=|S|-1}^{\infty} \gamma^k \delta \fV_T = \delta \gamma^{|S|-1} \fV_T.
        \end{aligned}
    \end{equation*}

    To satisfy inequality \eqref{eq::optimal_condition_inequality}, the following sufficient condition should hold:
    \begin{equation*}
        I^{\pi^+}_1 - I^{\pi}_2 \geq \frac{r_{\min} - r_{\max}}{1-\gamma} > I^{\pi}_3 \geq \delta \gamma^{|S|-1} \fV_T.
    \end{equation*}
    Rearranging the inequality gives the final result.
\end{proof}

\begin{theorem}[non-opposite optimal policies with negative rewards]
    \label{thm::condition4ONonworst}
    For a stochastic MDP with a discrete state space of size $|S| < \infty$, satisfying Assumptions \ref{assum::rewardBound} and \ref{assum::terminalAccessibility}, and $r_{\max} < 0$, if we set
    \begin{equation*}
        \fV_T > \frac{r_{\max} - r_{\min}}{\delta \gamma^{|S|-1}(1-\gamma)},
    \end{equation*}
    then $\pi^{+}_{\gamma} \notin \arg \min_{\pi} V^{\pi}$. This implies that the optimal policies of the two objectives are not opposite.
\end{theorem}

\begin{proof}
    The structure of this proof closely follows that of Theorem \ref{thm::condition4Optimal}, with the key difference being the signs of the rewards. Since \(r(s, a, s') < 0\) for all \((s, a, s') \in S^{\circ} \times A \times S\), the worst policy $\pi^-$ for the total return will prevents all non-terminal states \(s \in S^{\circ}\) from accessing any terminal state \(s' \in \partial S\),  as it results in \(V^{\pi}(s) = -\infty\) for all \(s \in S^{\circ}\). The existence of the policy is guaranteed by Assumption \ref{assum::terminalAccessibility} (2). 
    
    To ensure that \(\pi^+_{\gamma}\) is not the worst policy, we require that the discounted state values under \(\pi^+_{\gamma}\) exceeds the discounted state values of the worst policy \(\pi^-\) for all \(s \in S^{\circ}\). Following the analysis in Theorem \ref{thm::condition4Optimal}, the relation can be express by a similar inequality with three terms:  
    \begin{equation*}
        I^{\pi^-}_1 < I^{\pi^+_{\gamma}}_2 + I^{\pi^+_{\gamma}}_3,
    \end{equation*}
    where the terms \(I^{\pi^-}_1\), \(I^{\pi^+_{\gamma}}_2\), and \(I^{\pi^+_{\gamma}}_3\) are analogous to the definition \eqref{eq::rewardDefTerms} in Theorem \ref{thm::condition4Optimal} but are adapted for negative rewards. Using the bounds derived for \(I^{\pi^-}_1\), \(I^{\pi^+_{\gamma}}_2\), and \(I^{\pi^+_{\gamma}}_3\), we obtain the range for the terminal state value \(\fV_T\) as:  
    \begin{equation*}
        \fV_T > \frac{r_{\max} - r_{\min}}{\delta \gamma^{|S| - 1}(1 - \gamma)}.
    \end{equation*}
    
    When the terminal state value $\fV_T$ satisfies the condition above, the $\pi^+_{\gamma}$ is not the worst policy for total return, which means the optimal policies of the two objectives are not opposite, completing the proof.  
\end{proof}

\begin{remark}
    For the results in Theorems \ref{thm::condition4Optimal} and \ref{thm::condition4ONonworst}, if the MDP is deterministic, we have \(\delta = 1\), as stated in Remark \ref{remark::minAbsorptionProb}. Consequently, the calculation of the suitable range for the terminal state value can be simplified.
\end{remark}


Theorems \ref{thm::condition4Optimal} and \ref{thm::condition4ONonworst} provide the following insight: The terminal state value should be treated as a hyper-parameter when the objective misalignment problem arises. Specifically, \textbf{when the reward function is positive, the terminal state value should be reduced to encourage the agent to stay away from the terminal state. Conversely, when the reward function is negative, the terminal state value should be increased to encourage the agent to reach the terminal state as quickly as possible.}

Next, we present a special case of the two theorems above. This case represents a simple yet common setting: a deterministic MDP with a constant reward function (e.g. Grid World, CartPole, MountainCar, Acrobot). In this scenario, a suitable terminal state value ensures that the two objectives are fully aligned, while an unsuitable terminal state value makes the two objectives completely opposite. Furthermore, this result is not limited to discrete state spaces but extends naturally to countable state spaces.

\begin{theorem}[align with constant rewards]
    \label{thm::constantRewardOptimality}
    For a deterministic MDP with a countable state space \(S\), let the reward function be defined as: \(r(s, a, s') = r_C\) for all \((s, a, s') \in S^{\circ} \times A \times S\), and $r(s,a,s)=(1-\gamma)\fV_T$ for all $(s,a) \in \partial S \times A$, where \(r_C \in \mathbb{R}\). Then for any state $s \in S$, policies $\pi_1, \pi_2 \in \Pi$, and discount factor $\gamma \in (0,1)$, the following statements hold: 
    \begin{enumerate}[label = (\arabic*), itemindent = 0pt, labelindent = 0pt, labelwidth = 2em, labelsep = 5pt, leftmargin = *, itemsep=2pt, topsep=0pt, parsep=0pt]
        \item When \(r_C > 0\):  
        If \(\fV_T < \frac{r_C}{1 - \gamma}\), then \(V^{\pi_1}(s) > V^{\pi_2}(s)\) if and only if \(V^{\pi_1}_{\gamma}(s) > V^{\pi_2}_{\gamma}(s)\).  
        If \(\fV_T > \frac{r_C}{1 - \gamma}\), then \(V^{\pi_1}(s) > V^{\pi_2}(s)\) if and only if \(V^{\pi}_{\gamma}(s_1) < V^{\pi}_{\gamma}(s_2)\).  
        \item When \(r_C < 0\):  
        If \(\fV_T > \frac{r_C}{1 - \gamma}\), then \(V^{\pi}(s_1) > V^{\pi}(s_2)\) if and only if \(V^{\pi}_{\gamma}(s_1) > V^{\pi}_{\gamma}(s_2)\).  
        If \(\fV_T < \frac{r_C}{1 - \gamma}\), then \(V^{\pi}(s_1) > V^{\pi}(s_2)\) if and only if \(V^{\pi}_{\gamma}(s_1) > V^{\pi}_{\gamma}(s_2)\).  
    \end{enumerate}
\end{theorem}

\begin{proof}
    We provide the proof for the case $r_C > 0$ and $\fV_T < \frac{r_C}{1-\gamma}$. The proofs for other cases are similar.

    Proof of the ``if" part. We consider two trajectories $\tau^{\pi_1}(s)$ and $\tau^{\pi_2}(s)$ generated by policy $\pi_1, \pi_2 \in \Pi$, starting from $s \in S$, with lengths $m,n \in \mathbb{N}^+ \cup \{+\infty\}$, respectively. When $V^{\pi_1}(s) > V^{\pi_2}(s)$ holds, we clearly have $m > n$. The corresponding discounted values are given by:
    \begin{equation*}
        \begin{aligned}
            V_{\gamma}^{\pi_1}(s) & = \sum_{i=0}^{m-1} \gamma^i r_C + \sum_{i=m}^{\infty} \gamma^i (1-\gamma) \fV_T, \\
            V_{\gamma}^{\pi_2}(s) & = \sum_{i=0}^{n-1} \gamma^i r_C + \sum_{i=n}^{\infty} \gamma^i (1-\gamma) \fV_T.
        \end{aligned}
    \end{equation*}
    When $\fV_T < \frac{r_C}{1-\gamma}$ and $m>n$, we can easily verify that $V_{\gamma}^{\pi_1}(s) - V_{\gamma}^{\pi_2}(s) > 0$, completing the proof of the ``if" part.

    We prove the ``only if" part by contradiction. Let $\tau^{\pi_1}(s)$ and $\tau^{\pi_2}(s)$ be the same two trajectories we mentioned above. We assume $m < n$, and by solving $V_{\gamma}^{\pi_1}(s) - V_{\gamma}^{\pi_2}(s) > 0$, we obtain $\fV_T > \frac{r_C}{1-\gamma}$, which is contradict with the given condition $\fV_T < \frac{r_C}{1-\gamma}$. Hence, the assumption $m<n$ is false, implying $m \geq n$. Additionally, if $m=n$, we would have $V_{\gamma}^{\pi_1}(s) = V_{\gamma}^{\pi_2}(s)$, which also contradicts the condition $V_{\gamma}^{\pi_1}(s) > V_{\gamma}^{\pi_2}(s)$. Thus, we must have $m>n$, which further implies $V^{\pi_1}(s) > V^{\pi_2}(s)$. This completes the proof.
\end{proof}

\section{Objective Alignment Algorithm Design}
\label{sec::objective_alignment_algorithm}

In Section \ref{sec::policyAlignment}, we demonstrated that terminal state value modification enables optimal policies maximizing modified discounted returns to align with those maximizing total returns. However, this approach is constrained in its application scenarios. To address objective misalignment in more realistic DRL applications, we propose a trajectory reward data calibration method that ensures monotonic relationship between discounted returns and total returns.

Our method tackles two primary challenges: (1) estimating an appropriate discounted return for each trajectory, and (2) adjusting the current discounted return of a trajectory to match with the estimated value. To address the first challenge, we leverage historical trajectories collected from the environment as baselines for estimating suitable discounted returns. For the second challenge, we distribute the necessary adjustments across the trajectory’s reward data to enhance training stability.

The proposed method comprises three core components. The first component, Algorithm \ref{alg::estimate_hist}, estimates discounted returns for historical trajectories to establish a baseline for future estimation and offline training. The second component, Algorithm \ref{alg::estimate_new}, estimates suitable discounted returns for newly collected trajectories. The third component, Algorithm \ref{alg::adjust_reward}, modifies the reward data within a trajectory to match it with the estimated discounted return.

To implement these algorithms, an agent is initially trained for several steps to collect a few trajectories. These historical trajectories are processed using Algorithm \ref{alg::estimate_hist} to estimate their discounted returns, followed by Algorithm \ref{alg::adjust_reward} to adjust their reward data. The agent is then trained on these modified trajectory data in an off-policy manner, improving the utility of historical data and enhancing training stability. For newly sampled trajectories, Algorithms \ref{alg::estimate_new} and \ref{alg::adjust_reward} are applied to generate updated training trajectory data, supporting continued learning.

\begin{algorithm}  
\caption{Estimating the Suitable Discounted Return from Historical Data}  
\label{alg::estimate_hist}  
\begin{algorithmic}[1]  

\Require $n$ trajectories $\bm{\tau} = \{\tau_1, \dots, \tau_n\}$, discount factors $\gamma$, minimum slope $k_{\min}$, percentile $p$.
\Function{EstimateHist}{$\bm{\tau}$, $\gamma$, $k_{\min}$, $p$}   
    \For{each $i \in \{1, \dots, n\}$}
        \State Get total return $R(\tau_i)$ and discounted return $G(\tau_i)$. 
        \State Store $R(\tau_i)$ in $\fX$ and $G(\tau_i)$ in $\fY$.
    \EndFor  

    \State Sort $\bm{\tau}$ by $R(\tau_i)$ in ascending order as $\{\tau'_1, \dots, \tau'_n\}$.  
    \State Sort $\fY$ in ascending order as $\{G'_1, \dots, G'_n\}$.  
    \State Assign $G'_{i}$ to $\tau'_i$ to reorder discounted returns.

    \State $R^1_{\text{threshold}} = \min \fX + (\max \fX - \min \fX) \times p$.
    \State $R^2_{\text{threshold}}$ is the $p$ percentile of values in $\fX$.
    \State $R_{\text{threshold}} = \min(R^1_{\text{threshold}}, R^2_{\text{threshold}})$.

    \For{$i$ from $1$ to $n$}  
        \State Denote total return of $\tau'_i$ as $R'_i$.
        \If{$R'_i \geq R_{\text{threshold}}$ }  
            \State $k = \max(\frac{G'_{i} - G'_{i-1}}{R(\tau'_i) - R(\tau'_{i-1})}, k_{\min})$. 
            \State $ G^{\text{new}}_i = G^{\text{new}}_{i-1} + k \times (R(\tau'_i) - R(\tau'_{i-1}))$. 
        \Else
            \State $G^{\text{new}}_i = G'_{i}$.
        \EndIf  
    \EndFor  
  
    \Return $\{G^{\text{new}}_1, \dots, G^{\text{new}}_n\}$, $\fX$.  

\EndFunction  

\end{algorithmic}  
\end{algorithm}

We now delve into the details of the proposed algorithms. The core idea of Algorithm \ref{alg::estimate_hist} is to compute suitable discounted returns for all trajectories by reordering the discounted returns across trajectories and adjusting the slope between discounted returns and total returns. The algorithm requires four inputs: historical trajectories $\bm{\tau} = \{\tau_1, \tau_2, \dots, \tau_n\}$, a discount factor $\gamma$, and two parameters for slope adjustment—a minimum slope $k_{\min} \in (0,1)$ and a percentile $p \in (0,1)$. The algorithm consists of two main steps: (i) reordering discounted returns (lines 2–9) and (ii) slope adjustment (lines 10–21). The reordering step addresses the issue illustrated in Figure \ref{fig::objective_misalignment_problem} (a), where the relationship between total returns and discounted returns appears chaotic. The slope adjustment step addresses the phenomenon shown in Figure \ref{fig::objective_misalignment_problem} (b), where the slope between discounted returns and total returns is nearly zero.

In the first step, the total return $R(\tau_i)$ and discounted return $G(\tau_i)$ are calculated for each trajectory $\tau_i$. The trajectories $\bm{\tau}$ are then sorted in ascending order of their total returns, denoted as $\{\tau'_1, \dots, \tau'_n\}$, while the discounted returns $G(\tau_i)$ are similarly sorted in ascending order as $\{G'_1, \dots, G'_n\}$. Finally, the reordered discounted returns $G'_i$ are assigned to their corresponding sorted trajectories $\tau'_i$, ensuring that the discounted returns increase monotonically with the total returns.

In the second step, a slope adjustment threshold, $R_{\text{threshold}}$, is determined as the smaller value between the $p$-percentile of all total returns and the $p$-percentile of the range between the maximum and minimum total returns. This method ensures a reasonable threshold, particularly in cases where total returns are clustered. Since $\{\tau'_1, \dots, \tau'_n\}$ is sorted in ascending order, slope adjustment begins with $\tau'_1$. For a trajectory $\tau'_i$, if its total return $R(\tau'_i)$ is below $R_{\text{threshold}}$, the slope remains unchanged, and the corresponding discounted return is $G'_i$. Otherwise, the slope $k$ is computed as the maximum of its current slope, $(G'_i - G'_{i-1}) / (R(\tau'_i) - R(\tau'_{i-1}))$, and the minimum slope $k_{\min}$. This slope $k$ is then used to calculate the final adjusted discounted return $G^{\text{new}}_i$ for each trajectory $\tau'_i$. 

This algorithm outputs two sets: one containing the estimated discounted returns and the other containing the original total returns. These two sets are then used for estimating the discounted return for new trajectories.

\begin{figure}[h]
\centering
\includegraphics[width=0.45\textwidth]{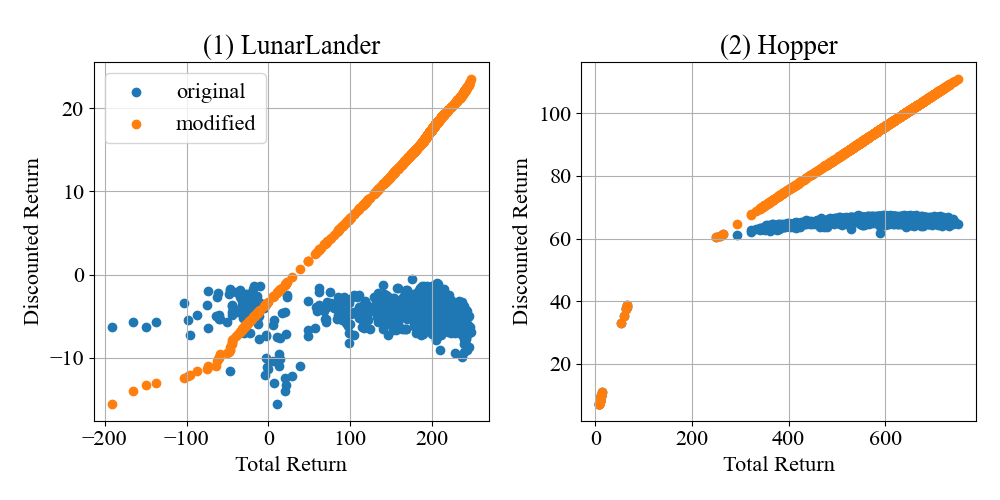}
\caption{
Demonstration of trajectories' original discounted returns and those modified by Algorithm \ref{alg::estimate_hist}. Each point represents a trajectory: blue points show the original total return (x-axis) and original discounted return (y-axis), while orange points represent the modified discounted return (y-axis) with the original total return preserved on the x-axis. The adjustment ensures that the modified discounted returns increase consistently with the original total returns.
}
\label{fig::adjusted_discounted_value}
\end{figure}

The Figure \ref{fig::adjusted_discounted_value} illustrates the processed results of historical trajectories after applying Algorithm \ref{alg::estimate_hist}. Each point represents a sampled trajectory. It can be observed that, after adjustment, the modified discounted return of each trajectory increases consistently with the original total return.

\begin{algorithm}  
\caption{Estimating the Suitable Discounted Return for a New Trajectory}  
\label{alg::estimate_new}  
\begin{algorithmic}[1]  
\Require current trajectory $\tau$, original total returns $\fX=\{R(\tau_1),...,R(\tau_m)\}$, discount factor $\gamma$, estimated discounted returns $\fZ=$ $\{G^{\text{new}}_1,$ $...,$ $G^{\text{new}}_m\}$
\Function{EstimateNew}{$\tau$, $\gamma$, $\fX$, $\fZ$, $k_{\min}$}  
    \State Get total return $R(\tau)$ and discounted return $G(\tau)$.
    \State Sort $\fX$ in ascending order as $\{R'_{1}, \dots, R'_{m}\}$.  
    \If{$R(\tau) \in [R'_i, R'_{i+1}], 0 \leq i < n$}  
        \State Obtain \( G^{\text{new}} \) by interpolating within \( [G^{\text{new}}_i, G^{\text{new}}_{i+1}] \).
    \ElsIf{$R < R'_1$}
        \State $G^{\text{new}} = G^{\text{new}}_1 - k_{\min} \times (R'_1 - R(\tau))$. 
    \ElsIf{$R > R'_n$}     
        \State $G^{\text{new}} = G^{\text{new}}_n + k_{\min} \times (R(\tau) - R'_n)$.
    \EndIf 
    
    \State store $G^{\text{new}}$ in $\fZ$, and $R(\tau)$ in $\fX$.  

    \Return $\fZ$, $\fX$.  

\EndFunction  

\end{algorithmic}  
\end{algorithm}

Algorithm \ref{alg::estimate_new} outlines the procedure for estimating the suitable discounted return for a new trajectory. This algorithm is executed after processing historical trajectories and is invoked each time a new trajectory is collected from the environment. The key idea is to estimate the discounted return by applying linear interpolation or extrapolation based on the estimated discounted returns of historical trajectories. The algorithm requires five inputs: the current trajectory $\tau$, the discount factor $\gamma$, the original total returns of all historical trajectories $\fX=\{R(\tau_1),...,R(\tau_m)\}$, the estimated discounted returns $\fZ=$ $\{G^{\text{new}}_1,$ $...,$ $G^{\text{new}}_m\}$, and the minimum slope $k_{\min}$. The process begins by calculating the total return $R(\tau)$ and discounted return $G(\tau)$ for the current trajectory $\tau$. If $R(\tau)$ falls within the range of total returns from the historical trajectories, $G^{\text{new}}$ is estimated using linear interpolation. Otherwise, extrapolation is performed using the minimum slope $k_{\min}$. Finally, the total return $R(\tau)$ of trajectory $\tau$ and its estimated discounted return $G^{\text{new}}$ are stored for future use.

\begin{algorithm}  
\caption{Adjust Reward Data Based on the Estimated Discounted Return}  
\label{alg::adjust_reward}  
\begin{algorithmic}[1]  

\Require trajectory $\tau$, estimated discounted return $D^{\text{new}}$
\Function{AdjustRewards}{$\tau$, $G^{\text{new}}$}
    \State Get the discounted return $G(\tau)$.
    \State Trajectory length of $\tau$ is $L$.
    \State $\Delta r = (G^{\text{new}} - G(\tau)) \times (1 - \gamma) / (1 - \gamma^{L - 1})$.

    \For{each $i \in \{1, \dots, n\}$}  
        \State Add $\Delta r$ on $r(s_i, a_i, s_{i+1})$. 
    \EndFor 

    \Return Trajectory $\tau^{\text{new}}$ with modified reward data.  

\EndFunction  

\end{algorithmic}  
\end{algorithm}  

Algorithm \ref{alg::adjust_reward} describes the procedure for adjusting rewards in a given trajectory after obtaining its estimated discounted return. The key idea is to distribute the difference between the estimated discounted return and the original discounted return evenly across all rewards in the trajectory. The algorithm takes two inputs: the current trajectory $\tau$ and the estimated discounted return $G^{\text{new}}$. It begins by calculating the discounted return $G(\tau)$. The reward adjustment $\Delta r$ is then computed as:
\begin{equation*}
    \Delta r =   \frac{(G^{\text{new}} - G(\tau)) \times (1 - \gamma)}{1 - \gamma^{L-1}}.
\end{equation*}
where $L$ represents the length of the trajectory. This adjustment $\Delta r$ is added to each reward in the trajectory. Finally, the algorithm returns the updated trajectory data.

\section{Experimental Results}

Building on the two methods presented for addressing the objectives misalignment problem, we now demonstrate their effectiveness through numerical experiments. First, we validate the correctness of the theorems in Section \ref{sec::policyAlignment}. Then, we present and analyze the experimental results of the proposed method outlined in Section \ref{sec::objective_alignment_algorithm}, focusing on settings with various discount factors and large trajectory length, to further highlight its effectiveness.

\subsection{Validation of the theorems for policy alignment}

\begin{figure*}[htbp]  
    \centering  
    \includegraphics[width=\textwidth]{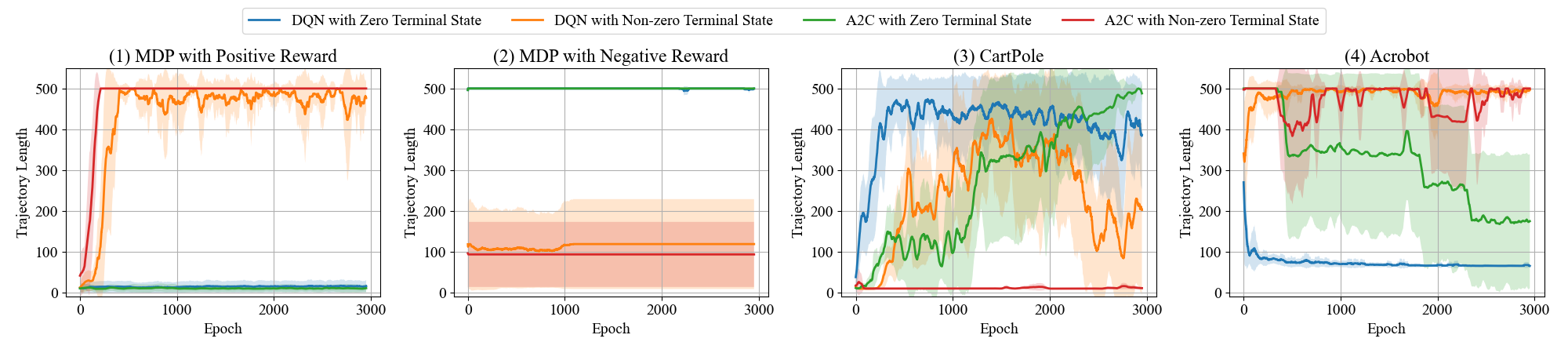} 
    \caption{
    Validation results for Theorems \ref{thm::condition4Optimal}, \ref{thm::condition4ONonworst}, and \ref{thm::constantRewardOptimality}. Solid lines indicate the mean trajectory length across five independent experiments (with different random seeds), smoothed using a window size of 50, and shaded regions show the standard deviations. The maximum trajectory length is setting to 500.
    In subfigure (1) (positive reward), a non-zero terminal state value yields the optimal policy with maximal trajectory length, substantiating Theorem \ref{thm::condition4Optimal}. Subfigure (2) (negative reward) demonstrates that a non-zero terminal state value generates a non-worst policy with trajectory length approximating 100, validating Theorem \ref{thm::condition4ONonworst}. Subfigure (3) ($+1$ reward) reveals that zero terminal state value escalates trajectory length to 500, signifying the optimal policy, whereas a non-zero value reduces trajectory length to zero, representing the worst policy, consistent with Theorem \ref{thm::constantRewardOptimality} (1). In subfigure (4) ($-1$ reward), zero terminal value diminishes trajectory length below 100, characterizing the optimal policy, while a non-zero value maintains it around 500, reflecting the worst policy, thus corroborating Theorem \ref{thm::constantRewardOptimality} (2).
    }  
    \label{fig::exact_result}  
\end{figure*}

To validate Theorems \ref{thm::condition4Optimal}, \ref{thm::condition4ONonworst}, and \ref{thm::constantRewardOptimality}, we design corresponding environments and conduct experiments using both value-based (DQN) and policy-gradient-based (A2C) methods. The following summarize the experimental setup, hyper-parameters, and results.

For both algorithms, training consists of $3 \times 10^3$ episodes with evaluations after each episode. The discount factor is $\gamma = 0.99$, and the maximum trajectory length is $500$. Both algorithms use three-layer fully connected networks with hidden layers of size $128$, optimized using the Adam optimizer. For DQN, parameter updates are performed at each sample, with a replay buffer size of $1 \times 10^6$, a batch size of $256$, a learning rate of $1 \times 10^{-3}$, and a target network updated via soft updates with $5 \times 10^{-3}$. For A2C, the learning rate is $1 \times 10^{-3}$, and the entropy parameter is set to $0.15$ for the Acrobot and $0.05$ for others. The results are demonstrated in Figure \ref{fig::exact_result}.


To validate Theorem \ref{thm::condition4Optimal}, we construct an MDP with a positive reward function. The MDP consists of ten states ($|S|=10$), including one terminal state, and three actions. The transition matrix is generated as follows: (1) a random matrix of size $9 \times 10 \times 3$ is sampled from a uniform distribution on $[1, 2]$, where indices $0 \sim 8$ represent non-terminal states and index $9$ represents the terminal state; (2) for each non-terminal state $i \in \{0, \dots, 8\}$, a random action $j \in \{0, 1, 2\}$ is chosen, and the element at $(i, 9, j)$ is set to zero to ensure a policy that avoids the terminal state; (3) the matrix is normalized to form a valid transition matrix. Rewards for transitions between non-terminal states are sampled from $[0.1, 0.2]$, while transitions to the terminal state are assigned a reward of $20$, resulting in $r_{\max} = 20$ and $r_{\min} = 0.1$. This reward structure increases the complexity of solving the MDP, highlighting the theorem's effectiveness. The results are shown in Figure \ref{fig::exact_result} (1). When the terminal state value is non-zero, it is set to $1.001 \times (r_{\min} - r_{\max}) / (\delta \gamma^9 (1-\gamma))$, where $\delta$ is derived from the transition matrix. Figure \ref{fig::exact_result} (1) validates the theorem by showing that non-zero terminal state values allow both A2C and DQN to avoid terminal states and maximize trajectory length, thereby representing the optimal policy. In contrast, setting the terminal state value to zero results in a worse policy.

To validate Theorem \ref{thm::condition4ONonworst}, we construct an MDP with a negative reward function. This MDP also consists of ten states ($|S|=10$), including one terminal state, and three actions. The transition matrix is generated as follows: (1) a random matrix of size $9 \times 10 \times 3$ is sampled from a uniform distribution on $[1, 2]$, where indices $0 \sim 8$ represent non-terminal states and index $9$ represents the terminal state; (2) a random non-terminal state $i \in \{0, \dots, 8\}$ and a random action $j \in \{0, 1, 2\}$ are selected, and only the element at $(i, 9, j)$ is retained, while all other elements with index $(\cdot, 9, \cdot)$ are set to zero, ensuring the terminal state $9$ is reachable only from state $i$ via action $j$; (3) the matrix is normalized to form a valid transition matrix. Rewards for transitions between non-terminal states are sampled from $[-0.2, -0.1]$, while transitions to the terminal state are assigned a reward of $-20$, resulting in $r_{\max} = -0.1$ and $r_{\min} = -20$. This design also increases the complexity of solving the MDP. The results are shown in Figure \ref{fig::exact_result} (2). When the terminal state value is non-zero, it is set to $1.001 \times (r_{\max} - r_{\min}) / (\delta \gamma^9 (1-\gamma))$. Figure \ref{fig::exact_result} (2) substantiates the theorem, showing non-zero terminal state values facilitate a non-worst policy with trajectories approximating 100 steps, while zero values lead to maximum trajectory lengths, indicating the worst policy.

These experiments demonstrate that setting the terminal state value to zero is bad in certain situations. Instead, setting it according to the theorems yields better results.

To validate Theorem \ref{thm::constantRewardOptimality}, we use the CartPole and Acrobot environments from Gym, assigning constant rewards of $r_C = +1$ and $r_C = -1$, respectively. Both environments feature continuous state spaces and discrete action spaces. The results are shown in Figures \ref{fig::exact_result} (3) and (4). For these experiments, when the terminal state value is non-zero, it is set to $1.001 \times r_C / (1-\gamma)$. Figures \ref{fig::exact_result} (3) and (4) validate the two statements.  In Figure \ref{fig::exact_result} (3), A2C achieves maximum trajectory length with zero terminal state value, signifying an optimal policy, whereas non-zero values result in premature termination within ten steps—representing the worst policy. DQN exhibits similar behavior, with non-zero values producing trajectories exceeding 100 yet still underperforming compared to zero-value settings. Figure \ref{fig::exact_result} (4) conclusively demonstrates that zero terminal state values allow both A2C and DQN to reach the terminal state within the maximum trajectory length, whereas non-zero values prevent reaching any terminal state, resulting in the worst policy.

\subsection{Performance of the objective alignment algorithm}

\begin{figure*}[htbp]  
    \centering  
    \includegraphics[width=\textwidth]{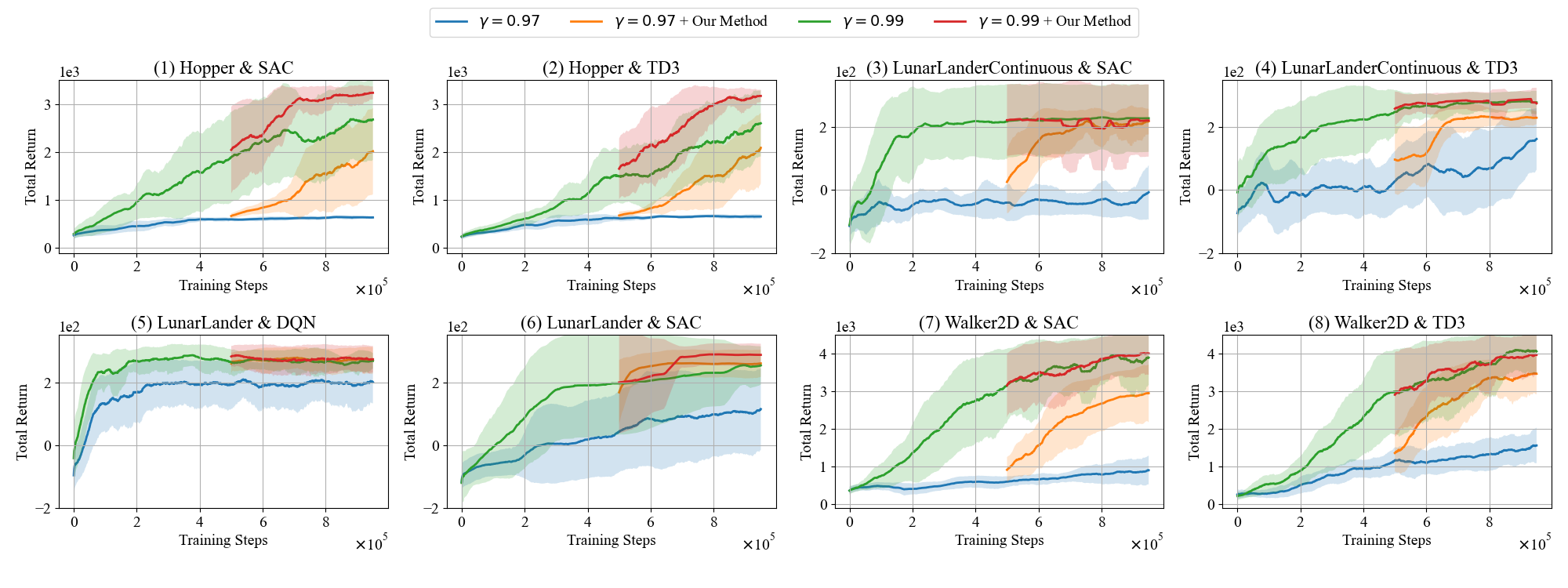} 
    \caption{
    Comparison of performance under different discount factors and their combination with our method. Solid lines show the mean performance over five independent experiments (with different random seeds), smoothed using a window size of 50, while shaded regions represent the corresponding standard deviations. 
    Our method significantly improves performance in all experiments with \(\gamma = 0.97\), particularly in the \textit{LunarLander} and \textit{LunarLanderContinuous} environments, where results are comparable to those with \(\gamma = 0.99\). Furthermore, in the \textit{Hopper} environment, our method achieves additional performance gains with \(\gamma = 0.99\).
    }  
    \label{fig::estimation_method_result}  
\end{figure*}





To ensure stability, our method relies on off-policy training. Therefore, we conduct experiments using three widely adopted off-policy algorithms: DQN, SAC (both discrete and continuous versions), and TD3. To demonstrate the generality of our proposed method, we test it on four commonly used control environments with continuous state spaces: \textit{LunarLander}, \textit{LunarLanderContinuous}, \textit{Hopper}, and \textit{Walker2D}. Notably, among these, only the first environment features a discrete action space.

The hyper-parameters used across all experiments are configured as follows. In all experiments, the total number of parameter updates is \( 1 \times 10^6 \), with our method starting at step \( 5 \times 10^5 \). After switch to our method, the algorithms train on modified historical trajectories for \( 6 \times 10^3 \) steps. Learned models are evaluated every \( 1 \times 10^3 \) training steps. All algorithms use a replay buffer size of \( 1 \times 10^3 \) trajectories, a batch size of $256$, and neural networks with three fully connected layers ($256$ hidden units, ReLU activations) optimized using Adam. Specifically, for DQN, the learning rate is \( 1 \times 10^{-3} \), parameters are updated every $20$ samples, and target networks are updated via soft updates with \(1 \times 10^{-3} \).  For discrete SAC, the learning rates for the critic and policy networks are \( 3 \times 10^{-4} \), the temperature factor learning rate is \( 1 \times 10^{-4} \), and the target entropy is set to \(-0.25\) times the action dimension. Parameters are updated every $20$ samples. For continuous SAC, the critic and policy learning rates are also \( 3 \times 10^{-4} \), with a temperature factor learning rate of \( 1 \times 10^{-4} \). The log standard deviation is constrained between $2$ and \(-20\), and the target entropy equals the action dimension. Parameters are updated at each sample.  
For TD3, the learning rates for the critic and policy networks are \( 3 \times 10^{-4} \). Exploration noise is sampled from a normal distribution with a standard deviation of $0.1$, while smooth noise is set to $0.2$ and clipped to \([-0.5, 0.5]\). The actor is updated every $2$ steps, with a random sample size of \( 1 \times 10^3 \) trajectories. Parameters are updated at each sample. Additionally, we incorporate the location data into the state representation for \textit{Hopper}, \textit{Walker2D}, and \textit{Humanoid}, thereby increasing the learning complexity and making the environments more representative of robotic control scenarios.

To demonstrate the effectiveness of our method in enhancing the robustness of off-policy algorithms to the discount factor, we first present experimental results using different discount factors. In this experiment, we set the percentile \( p = 1/3 \), with the minimum slope \( k_{\min} = 0.3 \) for \textit{LunarLander} and \( k_{\min} = 0.1 \) for other environments. The results are shown in Figure \ref{fig::estimation_method_result}. When the discount factor is small (\(\gamma = 0.97\)), transitioning to our method significantly improves performance across all environments by enhancing both the exploitation and exploration capabilities of the agent. For example, in Figure \ref{fig::estimation_method_result} (5) and (6) for the \textit{LunarLander} environment, the agent with \(\gamma = 0.97\) is capable of sampling high-return trajectories, as illustrated in Figure \ref{fig::objective_misalignment_problem} (1). However, it struggles to distinguish high-return policies from low-return ones. After applying our method, the agent's exploitation is improved, enabling it to better prioritize high-return trajectories. Moreover, in Figure \ref{fig::estimation_method_result} (1) and (2) for the \textit{Hopper} environment, our method improves the agent's exploration. Initially, the agent samples trajectories with total returns below $1000$, but by adjusting the slope between total and discounted returns, it begins to sample and effectively utilize higher-return trajectories, thereby strengthening its exploration capability. Additionally, with \(\gamma = 0.99\), our method effectively improves agent performance in the \textit{Hopper} environment, as shown in Figure \ref{fig::estimation_method_result} (1) and (2), demonstrating its robustness even with a large discount factor.

\begin{table}[htbp]
    \centering
    \begin{tabular}{cccc} 
         \hline 
         \textbf{Environment} & $\bm{k_{\min}}$ & \textbf{TD3} & \textbf{SAC} \\ 
         \hline
         \multirow{5}{*}{\textit{Walker2D}}
        
         & $0.1$ & $3470.3 \pm 547.9$ & $2955.5 \pm 758.8$  \\ 

         & $0.2$ & $3414.9 \pm 492.7$ & $2977.6 \pm 415.9$  \\

         & $0.3$ & $3413.9 \pm 323.9$ & $3003.7 \pm 475.8$  \\ 

         & $0.4$ & $3292.9 \pm 473.8$ & $2645.8 \pm 826.8$  \\

         & $0.5$ & $3040.9 \pm 718.9$ & $2692.1 \pm 362.3$  \\ 
        \hline
        \multirow{5}{*}{\parbox{2cm}{\centering \textit{LunarLander \\ Continuous}}} 
        
         & $0.1$ & $229.8 \pm 20.5$ & $218.3 \pm 39.5$  \\

         & $0.2$ & $209.0 \pm 77.9$ & $147.0 \pm 116.8$  \\ 

         & $0.3$ & $197.7 \pm 93.5$ & $149.7 \pm 102.5$  \\

         & $0.4$ & $189.1 \pm 96.1$ & $118.9 \pm 108.3$ \\

         & $0.5$ & $231.9 \pm 34.4$ & $163.5 \pm 89.9$  \\ 
         \hline
    \end{tabular}
    \caption{
    Performance comparison across different \(k_{\min}\). The experiments were conducted on \textit{Walker2D} and \textit{LunarLanderContinuous} with \(\gamma = 0.97\) and \(k_{\min}\) values ranging from 0.1 to 0.5. For \textit{Walker2D}, performance decreases as \(k_{\min}\) increases. In contrast, for \textit{LunarLanderContinuous}, a larger \(k_{\min} = 0.5\) still achieves good performance. This difference occurs because the maximum total return for \textit{Walker2D} is large, while that for \textit{LunarLanderContinuous} is small.
    }
    \label{tab::min_slope_diff}
\end{table}

The key hyper-parameter for our method is the minimum slope \( k_{\min} \). To assess its impact on performance, we conducted experiments with varying \( k_{\min} \) values in \textit{Walker2D} and \textit{LunarLanderContinuous} using TD3 and SAC with $\gamma=0.97$. As shown in Table \ref{tab::min_slope_diff}, for \textit{Walker2D}, performance declines when \( k_{\min} \) increases to 0.4 or 0.5. This is due to the high total return in \textit{Walker2D} (approximately 3500). With a large \( k_{\min} \), the modified discounted return may exceed 1000, making it difficult for the neural network to learn. In contrast, for \textit{LunarLanderContinuous}, where the total return is smaller (less than 300), a larger \( k_{\min} \) (e.g., 0.5 for TD3) still performs well. In such cases, a good strategy is to start with a small \( k_{\min} \) and gradually increase it. \textbf{Overall, when the total return is large, a smaller \( k_{\min} \) is preferable, but for smaller total returns, trying different \( k_{\min} \) values beginning with a small one is recommended.}

\begin{figure}[h]
\centering
\includegraphics[width=0.45\textwidth]{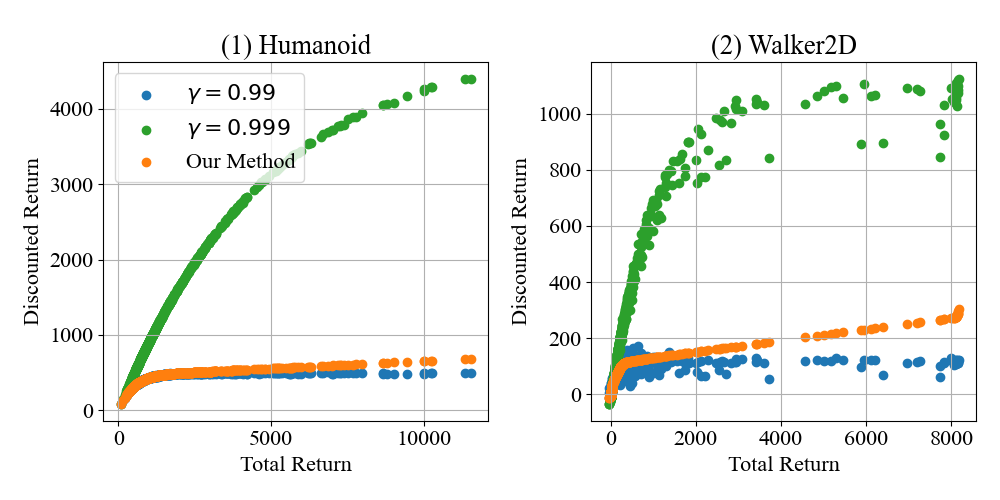}
\caption{
Demonstration of trajectories' discounted returns with different discount factors. Each dot represents a trajectory, and the orange dots are computed using Algorithm \ref{alg::estimate_hist} with parameters \(k_{\min}=0.02\), \(p=1/3\), and \(\gamma=0.99\). For \(\gamma=0.99\), the discounted return remains manageable (e.g., below 1000) but becomes indistinguishable when the total return exceeds 2000. In contrast, for \(\gamma=0.999\), the slope is steeper, but the discounted return quickly become too large (e.g., exceeds 1000). By comparison, our method adjusts the slope effectively and simultaneously controls the maximum value of the discounted return.
}
\label{fig::long_adjusted_discounted_value}
\end{figure}

Our method is well-suited for environments with large trajectory lengths. In such cases, the discount factor often needs to be increased to ensure that the discounted return increases monotonically with the total return. However, larger discount factors can result in excessively large discounted returns, as shown for \(\gamma = 0.999\) in Figure \ref{fig::long_adjusted_discounted_value}, making it challenging for the neural network to accurately fit these values. Our method addresses this issue by controlling the magnitude of the discounted returns using a smaller discount factor and \( k_{\min} \), effectively mitigating the problem. 

\begin{figure*}[htbp]  
    \centering  
    \includegraphics[width=\textwidth]{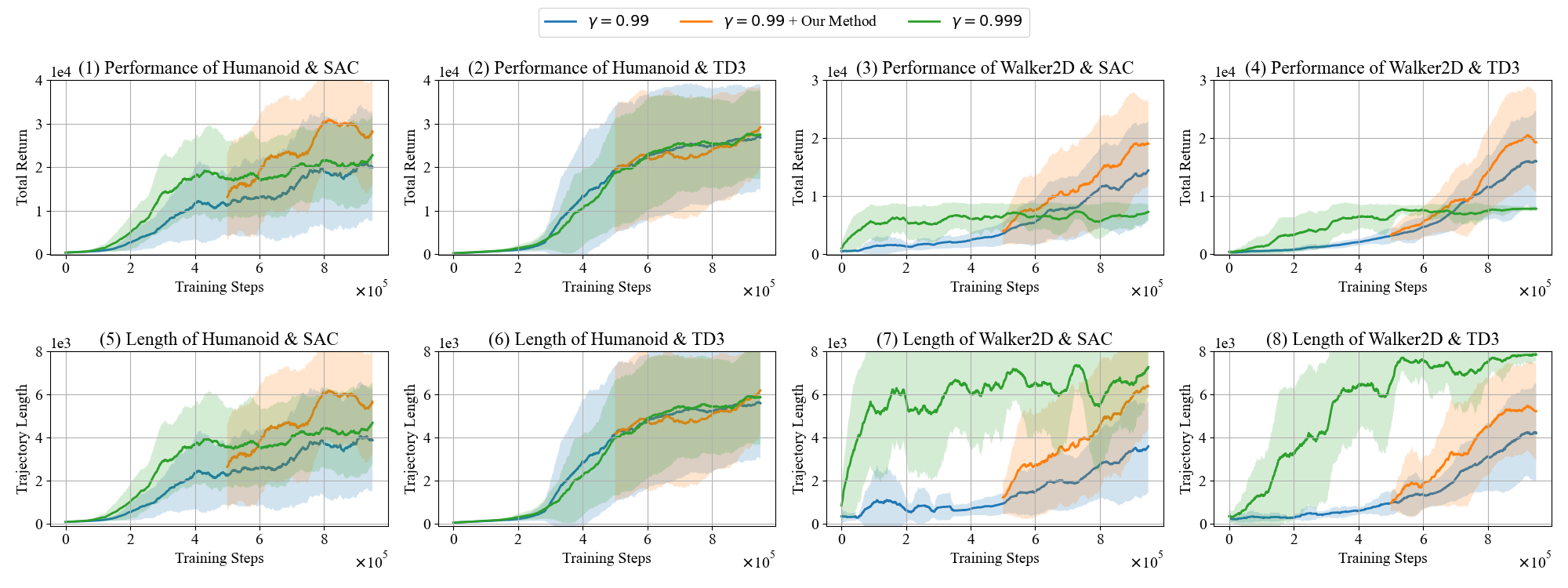} 
    \caption{
    Comparison of performance and trajectory length across various discount factors and their combinations with our method. Solid lines represent mean values from five independent experiments (each with different random seeds), smoothed using a window size of 50, while shaded regions indicate the corresponding standard deviations. Subfigures (1) to (4) illustrate performance, and (5) to (6) depict trajectory length. Our method for TD3 on \textit{Humanoid} outperforms both the \(\gamma=0.99\) and \(\gamma=0.999\) settings at the final training stage, with significant performance enhancements observed in other scenarios as well. This improvement is attributed to a increase in trajectory length, as evidenced in (5) to (8).
    }  
    \label{fig::long_estimation_method_result}  
\end{figure*}

To evaluate our approach, we conducted experiments in environments with large trajectory lengths, including \textit{Humanoid} and \textit{Walker2D}, with a maximum trajectory length of 8000. In this experiment, we set \( p = 1/3 \), \( k_{\min} = 0.02 \) for SAC, \( k_{\min} = 0.01 \) for TD3 in \textit{Walker2D}, and \( k_{\min} = 0.0075 \) for TD3 in \textit{Humanoid}. The results are presented in Figure \ref{fig::long_estimation_method_result}.

Figure \ref{fig::long_estimation_method_result} (1) to (4) compare performance across different settings. For \textit{Humanoid} with TD3, our method outperforms both the $\gamma = 0.99$ and
$\gamma = 0.999$ settings at the final training stage. For other scenarios, our method significantly improves performance for \(\gamma = 0.99\) and surpasses the performance achieved with \(\gamma = 0.999\). This improvement arises from our method's ability to effectively increase the trajectory length, as demonstrated in Figure \ref{fig::long_estimation_method_result} (5) to (8). However, simply increasing the trajectory length does not necessarily enhance performance. For example, as seen in Figure \ref{fig::long_estimation_method_result} (7) and (8), the results with \(\gamma = 0.999\) have larger trajectory lengths, yet the performance is poor. This occurs because \(\gamma = 0.999\) results in excessively large discounted returns.

As shown in Figure \ref{fig::long_adjusted_discounted_value}, the discounted return can easily exceed 1000, making it difficult for the neural network to accurately fit the discounted state values or learn their correct rank. This adversely affects the policy network's ability to optimize, ultimately leading to poor performance. In contrast, our method controls the value of the discounted return by adjusting the slope size, facilitating more accurate estimation of the discounted state values. With this improved value estimation, our method effectively increases the ``useful" trajectory length, leading to substantial performance improvements.

The experimental results above demonstrate that ensuring the discounted returns to increase monotonically with the total returns at the trajectory level can effectively improve performance. Notably, in our method—particularly in Algorithm \ref{alg::adjust_reward}—the reward data is rescaled and its sign altered, stripping it of its original physical meaning. Despite this, performance still improves. This observation leads to the key insight that \textbf{maintaining alignment between objectives is more crucial than preserving the original scale and sign of the reward data. This insight paves the way for developing new proxies that naturally consistent with the total return.}

\section{Conclusion and Future Work}

In this work, we identified a fundamental yet often overlooked problem in RL: objective misalignment, where the policy maximizing discounted returns fails to maximize total returns. Specifically, in practical applications, this issue arises from the fact that the discounted returns of trajectories do not monotonically increase with their total returns. Our analysis of the performance gap between the policy that maximizes discounted returns and the one that maximizes total returns reveals that simply increasing the discount factor is insufficient, especially in environments with cyclic states, which are common in RL. To address this problem, we proposed two alternative approaches. First, we introduced the terminal state value as a tunable hyper-parameter to align policies that optimize both discounted and total returns. Through theoretical analysis, we identified suitable ranges for this hyper-parameter, challenging the conventional assumption that terminal state values must always be set to zero. Second, for practical DRL applications, we developed a trajectory reward data calibration method that adjusts sampled trajectory rewards to achieve alignment at the trajectory level. This method enhances the robustness of off-policy RL algorithms to the discount factor and improves performance in tasks with long trajectories. Our results highlight the potential of reward data calibration techniques to address objective misalignment and suggest new directions for designing alternative proxies to the total return beyond the discounted return.

From the above discussion, we identify objective misalignment as a fundamental issue affecting the performance of RL algorithms. However, our reward data calibration method addresses this problem only at the trajectory level and in an off-policy manner. A more fundamental solution would involve designing a new proxy that inherently increases with the total return while ensuring its maximum value remains controllable. We believe the insights gained from this work can inspire the development of such proxies with improved properties, ultimately resolving the misalignment issue more effectively.

\section*{Acknowledgment}
This work is sponsored by the STI 2030—Major Projects 2022ZD0208700(S. Y., F. W., P. L.), Shanghai Municipal of Science and Technology Major Project No. 2021SHZDZX0102.(S. Y., F. W., P. L., T. L.), MoE Key Lab of Artificial Intelligence, AI Institute, Shanghai Jiao Tong University, China(S. Y., F. W., P. L.), the National Key R\&D Program of China Grant No. 2022YFA1008200 (T. L.), the National Natural Science Foundation of China Grant No. 12101401 (T. L.), Shanghai Municipal Science and Technology Key Project No. 22JC1401500 (T. L.),

\bibliographystyle{IEEEtran}
\bibliography{references}
\end{document}